\documentclass{article}

\usepackage[numbers]{natbib}
\PassOptionsToPackage{numbers, compress}{natbib}

\usepackage[final]{neurips_2023} 

\usepackage[utf8]{inputenc} 
\usepackage[T1]{fontenc}    
\usepackage{hyperref}       
\usepackage{url}            
\usepackage{booktabs}       
\usepackage{amsfonts}       
\usepackage{nicefrac}       
\usepackage{microtype}      
\usepackage{xcolor}         
\usepackage{algorithm}
\usepackage{algorithmic}
\usepackage{graphicx,amsmath,amssymb,tikz,psfrag,mathtools}
\usepackage{makecell}

\title{Seller-side Outcome Fairness in Online Marketplaces}

\author{%
Zikun Ye$^{1}$ \quad Reza Yousefi Maragheh$^{2}$ \quad Lalitesh Morishetti$^2$ \quad Shanu Vashishtha$^2$  \\
\textbf{Jason Cho}$^2$ \quad \textbf{Kaushiki Nag}$^2$ \quad \textbf{Sushant Kumar}$^2$ \quad \textbf{Kannan Achan}$^2$ \\
$^1$University of Washington \quad $^2$Walmart Global Tech\\
}

\begin{document}

\maketitle

\begin{abstract}
  This paper aims to investigate and achieve seller-side fairness within online marketplaces, where many sellers and their items are not sufficiently exposed to customers in an e-commerce platform. This phenomenon raises concerns regarding the potential loss of revenue associated with less exposed items as well as less marketplace diversity. We introduce the notion of seller-side outcome fairness and build an optimization model to balance collected recommendation rewards and the fairness metric. We then propose a gradient-based data-driven algorithm based on the duality and bandit theory. Our numerical experiments on real e-commerce data sets show that our algorithm can lift seller fairness measures while not hurting metrics like collected Gross Merchandise Value (GMV) and total purchases.
\end{abstract}

\section{Introduction}
In e-commerce settings, a fair recommendation system ensures that all users, items, and sellers have equal opportunities to be represented and recommended \cite{rastegarpanah2019fighting}. This is important from two aspects. (i) User satisfaction: A fair recommendation system ensures that users are exposed to diverse products, reducing the likelihood of echo chambers and promoting serendipitous discoveries \cite{chaney2018algorithmic}. (ii) Business sustainability: Fairness in recommendations helps to maintain a healthy ecosystem of sellers and service providers in online marketplace platforms, fosters competition and innovation, and is in line with the long-term interest of recommendation systems \cite{abdollahpouri2019managing}. 

Recent studies have begun to explore fairness in e-commerce recommendation systems. However, most of these investigations target the user side \cite{li2021user, pitoura2022fairness}, leaving seller-side fairness relatively unexplored. This lack of focus on seller-side fairness can lead to various issues, including popularity bias, where popular items are overly exposed in recommendations at the cost of less popular items that customers may find interesting.

Our analysis of real data sets shows that a small percentage of popular third-party online sellers attain the majority of clicks, indicating a significant imbalance in the distribution of exposure and opportunities. This skewed exposure can harm the overall e-retail ecosystem, as it stifles the growth of smaller sellers and reduces the diversity of products available to consumers. When the recommendation system predominantly promotes popular items or those from prominent sellers, it homogenizes the available options, which in return, can lead to a decrease in overall consumer satisfaction and utility \cite{chaney2018algorithmic}. 

Another critical drawback of the lack of seller-side fairness is the potential loss of seller loyalty. When sellers perceive that their products are not being fairly promoted or given equal opportunities for sales on a platform, they may become unsatisfied with the promotion performance in the marketplace. This dissatisfaction can lead to sellers seeking alternative platforms that offer greater visibility and fairer promotion of their products. The exodus of sellers would decrease the platform's inventory and harm its reputation, further diminishing its attractiveness to both new sellers and consumers \cite{ye2023cold}. A similar issue can happen in the online advertisement when the publisher does not expose the ads to a sequence of users arriving at the platform \cite{feldman2010online}.

Given these drawbacks, our research aims to design a data-driven algorithm that addresses the popularity bias issue while achieving diversity and outcome fairness in recommendation systems. By ensuring fairness in the allocation of exposure and sales opportunities, we can create a more equitable marketplace that benefits both sellers and consumers. However, we also recognize the potential short-term loss of revenue that may result from this increased outcome fairness, as users may initially be less familiar with less popular or niche products. Our goal is to balance the trade-off between promoting outcome fairness and maintaining platform revenue. Based on duality theory, our algorithm leverages stochastic gradient descent for the dual model. A similar dual mirror descent algorithm is proposed for online allocation problems \cite{balseiro2023best}. In addition to the exploitation, our algorithm also incorporates a new exploration scheme called the ``inverse proportional to the gap'' rule \cite{simchi2022bypassing}.

We empirically evaluate our algorithm by doing extensive experiments on proprietary data containing more than 600 million data points and more than 40 thousand sellers. We have three evaluation criteria: revenue (or GMV), outcome fairness, and total long-term/objective value.

\section{Related Work}
Our paper is primarily related to three streams of literature: online allocation, multi-armed bandit, and fairness in allocation.

The product recommendation process can be viewed as an online allocation problem, widely studied in computer science and operations research. For example, authors in Balseiro et al. \cite{balseiro2015repeated} have developed algorithms to solve these problems in the context of online advertising with budget constraints. Classic literature in this field focuses on linear reward functions \cite{mehta2007adwords,devanur2009adwords}. Our study extends this by introducing a non-separable fairness term in the objective function, thus enriching the existing literature.

Less exposed items pose the challenge of inaccurate reward function prediction in our e-commerce setting. The prediction is made through complex models, such as neural networks, tree-based models, etc. This leads us to multi-armed bandit literature, aiming for interactive learning of reward functions online. However, widely studied exploration schemes, including $\epsilon$-greedy, upper confident bound \cite{chu2011contextual}, and Thompson sampling \cite{agrawal2013thompson} for the linear contextual bandit, fail to work for general prediction models. We refer interested readers to \cite{ye2023cold,simchi2022bypassing} for a more comprehensive discussion and literature review. We address this challenge by using a novel ``inverse proportional to the gap'' exploration scheme \cite{simchi2022bypassing}.

Our research also connects to the growing fairness literature. In particular, fairness has emerged as a critical aspect of user satisfaction and trust, and it can influence user engagement, retention, and long-term platform success \cite{kamishima2012fairness, yao2017beyond, farnadi2018fairness}. The growing focus on fairness in recommendation systems can be observed through numerous recent studies that explore various aspects of this topic, indicating that fairness is indeed an active research area. Kamishima et al. \cite{kamishima2012fairness} present a fairness-aware machine learning approach to recommendation systems, addressing the need for fair recommendations by minimizing discrimination based on user attributes. Yao and Huang \cite{yao2017beyond} examine the trade-offs between recommendation accuracy and fairness, proposing a framework for achieving fairness in collaborative filtering while maintaining high levels of accuracy. Farnadi et al. \cite{farnadi2018fairness} introduce a fairness-aware ranking algorithm that explicitly considers the diverse interests of users and seeks to promote fairness in personalized recommendations. 

Various fairness metrics have been proposed in the literature to quantify and assess the degree of fairness in recommendation systems. These metrics include but are not limited to demographic parity, disparate impact, and equal opportunity \cite{hardt2016equality, beutel2017data, zehlike2017fa}. Each of these metrics captures different aspects of fairness and can be applied in various contexts depending on the specific fairness objectives of a recommendation system. This multitude of fairness measures indicates that there is no one-size-fits-all approach to fairness, and different measures may be more suitable for different applications. In this paper, we specifically focus on seller-side fairness, examining the distribution of exposure and opportunities among sellers in e-commerce platforms. While the majority of existing fairness literature concentrates on user-side fairness, there is a growing interest in understanding and addressing seller-side fairness in recommendation systems. This aspect of fairness is particularly relevant for e-commerce platforms, where seller-side biases can negatively impact the diversity and competitiveness of the market, leading to reduced user satisfaction and suboptimal economic outcomes. For example, Abdollahpouri et al. \cite{abdollahpouri2017controlling} propose a method for controlling the exposure of items in recommendation systems, which can help balance the exposure between popular and less popular items. Biega et al. \cite{biega2018equity} propose a framework for achieving equity of attention in recommendation systems, focusing on fairly distributing the visibility of items among various groups. In a related paper, Morik et al. \cite{morik2020controlling} address a similar seller-side fairness problem. However, their algorithm is based on a merit-based exposure-allocation criterion, which prioritizes the merit and relevance of items rather than solely based on user engagement metrics. Our approach does not need inputs of ``merit'' for ranking; instead, we are more focused on the explicit and aggregated outcome of sellers to achieve outcome fairness.

\section{Problem Formulation and Algorithm}

We start by describing our online allocation problem. At each time when customer view comes, indexed by $t$, we have to decide $K$ different products to display on one page with the side information, e.g., (predicted) Conversion Rate (CVR) or Click-Through-Rate (CTR) and price information. We formulate the problem as the following general optimization model with total $T$ time periods (i.e., exposures, customer views), indexed by $t \in [T]$, and $m$ sellers, indexed by $j \in [m]$. We assume that each seller $j$ has $K_j$ different products to be sold. 

\begin{equation}\label{problem:primal-model}
\begin{aligned}
\max_{x}~&\sum_{t=1}^T\sum_{j=1}^m\sum_{k=1}^{K_j} p_{jk}\textcolor{black}{c_{tjk}}x_{tjk}+r(\cdot)\\
\textit{s.t.}~&\sum_{j=1}^m \sum_{k =1}^{K_j} x_{tjk}\leq K,~\forall t\in[T]\\
&x_{tjk}\in [0,1],~\forall t\in[T],~j\in[m],~k\in[K_j],
\end{aligned}
\end{equation}
where $x_{tjk}$ is decision variable representing the probability of displaying product $k$ of seller $j$ to customer $t$. One can also view $x_{tjk}$ as a relaxation of the binary decision on whether the item is displayed. This linear relaxation only incurs a very small optimization error under a very mild assumption \cite{ye2023cold}. $c_{tjk}$ is the CVR (or CTR) for each item $k \in [K_j]$ from seller $j \in [m]$ at time $t$. We define $p_{jk}$ as the unit revenue for the product $k \in [K_j]$ from seller $j \in [m]$. So, the first term in the objective function is the expected generated GMV from the recommended set of items. Note that in case optimizing on the total clicks or conversions instead of GMV, one can replace $p_{jk} =1$, giving the same weight for each item. The first constraint represents that $K$-cardinality cap on the number of recommended items presented in a marketplace page/recommendation module. Note that budget and resource consumption constraints can also be readily added to our model. But for ease of exposition, we mainly focus on the fairness part in the objective function and use the notation $x\in\mathcal{X}$ to represent convex constraints in the following. Also, note that our linear revenue function in the objective can be extended to a general concave function; see the detailed discussion in \cite{balseiro2023best}.

The second term, $r(\cdot)$, is a multi-dimensional concave function in aggregated outcomes (total exposures, clicks, conversions, GMV) at the aggregated seller or item level. In the following, to be consistent with the business practice, we focus on the seller-level outcome, i.e.,  $r(a)=r(a_1,a_2,...,a_m)$ where $a_j$ represents the aggregated outcome of seller $j\in[m]$. We call it a regularization term, which can be quite flexible, like any of the following examples.

\begin{itemize}
\item []\textbf{{Example 1.}} \textit{No Regularizer}: $r(a)=0$. In this case, we merely maximize the GMV (or CTR) metric without any regularization.

\item []\textbf{{Example 2.}} \textit{Above-Target Revenue Regularizer}: $r(a)= \sum_{j=1}^m \beta_j \min(a_j,\alpha_j)$ where $a_j=\sum_{t=1}^T \sum_{k=1}^{K_j}p_{jk}c_{tjk}x_{tjk}$ denotes the collected GMV by seller $j$, with the threshold $\alpha_j\geq 0$. This regularizer can be used when the marketplace platform wants the sellers to attain the expected target GMV of $\alpha_j$ or more. $\beta_j$ is the implicit penalty rate for the marketplace platform when seller $j$ collects less GMV than target $\alpha_j$. 

\item []\textbf{{Example 3.}} \textit{Max-min Revenue Fairness Regularizer}: $r(a)=\beta \min_{j\in[m]}a_j$ where $a_j$ is defined same as Example 2. This regularizer imposes outcome fairness for sellers, i.e., no seller gets too little GMV.
\end{itemize}

The first example covers the regular decision problems without regularizers. The second fairness regularizers fit well in applications where there are certain consumption goals. For example, in digital advertising, advertisers prefer their budgets to be spent as much and quickly as possible. In the remaining paper, we choose this regularize for our application. Because we find that too little GMV or low exposure can decrease sellers' satisfaction and loyalty, making them switch to other platforms for their business, which is consistent with the empirical evidence in previous literature \cite{ye2023cold}. Essentially, the long-term benefit is not linear in the scale of short-term revenue, and this regularizer assigns higher value to those low-exposure sellers. The third max-min fairness notion seeks to maximize the minimum consumption across sellers and has been extensively in the literature \cite{bertsimas2011price, bertsimas2012efficiency}. However, this max-min fairness notion may be too pessimistic for our application since the worst seller performance is typically zero revenue. Note that all the above fairness notions are outcome-oriented in contrast to those merit-based exposure-allocation criteria \cite{morik2020controlling}, which rely on the items' intrinsic merits for adjusting ranking decisions. However, our framework can also easily capture the merits of items by simply taking $c_{tjk}$ as a combination of CTR/CVR and merit score. From the seller side, our second fairness notion can flexibly assign heterogeneous values $\alpha_j$ and $\beta_j$ across sellers to capture the merits of sellers.

\subsection{The Dual Model}
Primal decision space in model \eqref{problem:primal-model} is large. Thus, it is natural to turn to the dual decision space. Using the results in a dual-descent algorithm and an easy-to-implement ranking policy. As we discuss above, we use the notation $x\in\mathcal{X}$ to represent constraints. In the following, our discussion focuses on the primal model with \textit{Above-Target (Seller level) Purchases} regularization term, i.e., we want to let each seller $j$ have more than  {$\alpha_j$} purchases over all their products and $\beta_j$ is the weight for the regularization term. For ease of exposition and illustration, we set the cardinality cap $K=1$ in the remaining paper. But note that our model and algorithm are easily adapted to any $K\in\mathbb{Z}^+$. Under the above specification, our problem can be formulated as the model \eqref{primal},

\begin{equation}\label{primal}
\max_{x\in\mathcal{X}}~\sum_{t=1}^T \sum_{j=1}^m\sum_{k=1}^{K_j} p_{jk}c_{tjk}x_{tjk}+\sum_{j=1}^m \beta_j \min \{ \sum_{t=1}^T\sum_{k=1}^{K_j}x_{tjk}c_{tjk},\alpha_j \}.
\end{equation}

We derive the equivalent Lagrangian dual model \eqref{dual} in the following, 

\begin{equation}\label{dual}
    \min_{ {0\leq\lambda_j\leq\beta_j}}~\sum_{t=1}^T\max_{x\in\mathcal{X}}\{ (p_{jk}+ {\lambda_j})c_{tjk}x_{tjk}\}-\sum_{j=1}^m \alpha_j {\lambda_j},
\end{equation}
where the decision variables $\lambda_j$ are constrained in intervals $[0, \beta_j]$ for all $j\in[m]$, and the inner maximization problem can be easily solved period by period.

\begin{algorithm}[b]
\caption{Online Dual Gradient Descent with Learning}
	\begin{algorithmic}[1]\label{alg1}
		\STATE \textbf{Initialize:} decision variables ${\lambda_j^1=0}~\forall j\in[m]$, predicting machine learning model $M^1$ and its training method $\mathcal{L}$, step size $\eta$, learning rate $\gamma_t$, the total number of items $H$\\
		\FOR{$t=1$ to $T$ \do} 
		\STATE {\bf Step 1:} Obverse the realization of contextual information $s_{tjk}$ for prediction. Predict CVR $\hat{c}_{tjk}$ via the model $M^t(s_{tjk})$, for all sellers and their products.\\
            \STATE {\bf Step 2:} Compute the fairness-aware rank score $\hat{f}_{tjk} = (p_{jk}+{\lambda_j^t}) \hat{c}_{tjk}$ for all sellers $j\in[m]$ and products $k\in[K_j]$. Then, find the empirical optimal product $\hat{l}$ with the highest score, i.e., $\hat{l} = \textit{argmax}_{j\in[m], k\in[K_j]} (p_{jk}+{\lambda_j^t}) \hat{c}_{tjk}$.\\
            \STATE {\bf Step 3:} (Exploitation-Exploration trade-off)\\
                    \STATE ~~~~~ With probability $\frac{1}{H+\gamma_t(\hat{f}_{t\hat{l}}-\hat{f}_{tjk})}$ to randomly display the other item $jk$. Otherwise, choose the empirical optimal item $\hat{l}$. \\
		\STATE {\bf Step 4:} Observe outcomes and update dual variables via projected subgradient descent, ${\lambda_j^{t+1}}=\text{Proj}_{[0,\beta_j]}\big(\lambda_j^{t}-\eta(\sum_{k=1}^{K_j}\mathbb{I}_{\text{item } {jk} \text{ is purchased}}-\alpha_j/T)\big)$.\\
        \STATE {\bf Step 5:} Update the prediction model $M^{t+1}=\mathcal{L}(M^{t})$.
		\ENDFOR
	\end{algorithmic}
\end{algorithm}

\subsection{Algorithm}
Algorithm \ref{alg1} presents the main algorithm we propose in this paper. Note that without the exploration, i.e., Step 3, the algorithm is a standard dual gradient descent algorithm from online convex optimization \cite{hazan2016introduction}, and the theoretical performance is well studied \cite{balseiro2021regularized}. In other words, if we have an unbiased estimation of the CVR (or CTR) ${c}_{tjk}$, the algorithm without Step 3 attains the optimal sub-linear regret compared to the hindsight solution \cite{balseiro2021regularized}. 

However, in practical settings, the predicted purchase rates for those less-exposed items are inaccurate. This means we need to actively and smartly explore those less-exposed items to gain higher potential GMV (revenue). To deal with this, our algorithm incorporates the ``inverse proportional to the gap'' \cite{simchi2022bypassing} exploration rule. Also, we use the notation $\hat{c}_{tjk}$ to present the predicted purchase rate by the utilized ML model.

At each period, our algorithm first predicts the purchase rate in Step 1. In Step 2, the algorithm finds the recommendation set. Step 3 leverages the ``inverse proportional to the gap'' sampling rule to balance the exploitation and exploration. In Steps 4-5, we update the dual decision variables and predicting models. Note that the predicting model can be updated periodically instead of in each period, and the learning rate should be adjusted accordingly (see \cite{simchi2022bypassing}).

\section{Numerical Experiments}
In this section, we present numerical experiments based on two data sets obtained from real industrial online marketplace platforms. Third-party sellers can sell their items on each one of these platforms. We used a \textit{Above-Target (Seller level) Clicks} fairness regularizer in these experiments. 

\subsection{Data and Implementation Details}
We evaluate our proposed recommendation algorithm on two simulations based on two datasets: 
\begin{enumerate}
    \item Proprietary Dataset: A proprietary user user-item interaction data set randomly sampled from a recommendation module of an online marketplace platform. In the dataset, each data point contains information about the customer, anchor item, recommendation items, and whether those items are clicked and purchased or not. The dataset is sampled from the marketplace interactions and includes more than 600 million data points, including a large number of sellers. To evaluate our algorithm on this dataset, we create a deep neural network-based simulator to simulate the online environment and then run the algorithm against the simulator. This method of building a simulator for evaluating bandit algorithms is widely used \cite{chapelle2011empirical}.
    \item Electronics Event History (EVS) Dataset \cite{kaggle}: The dataset is publicly available and includes  36,966 items from  999 brands and feedback data of 490,399 user sessions. Using this data, we estimate the aggregate CVR and the associated price of brands as the average of experimental CVRs of the items of the brand. Then, using these representative items for each brand, we simulate online data. Due to data sparsity and in order to have a more robust simulation with a smaller variance base for the CVR of items, we filter out all the brands with less than $10,000$ item views.
\end{enumerate}

In our numerical experiments, we use the regularization term by \textit{Above-Target (Seller level) Clicks}, i.e., $r(a)=\sum_{j=1}^m \beta_j \min(a_j,\alpha_j)$ where $a_j=\sum_{t=1}^{T}\sum_{k=1}^{K_j}c_{tjk}x_{tjk}$, and $c_{tjk}$ is CTR. 

For proprietary data, we aim to achieve a fair outcome with more than $5$ clicks for each seller per a $T$ visit (which we mask due to proprietary agreement), i.e., we set $\alpha_j=\alpha=5$ per $T$ for any seller $j\in [m]$.  We choose $\alpha_j=\alpha =5$ as per $T$ visits since approximately 80\% of sellers are observed to get less clicks than the threshold. Our objective is to see the effect of the proposed algorithm on the number of sellers attaining more clicks. For the EVS public data set, we set this value to be a $5$ transaction for each brand for every $200,000$ visit. We test out different values $\beta_j$ in numerical experiments to illustrate the trade-off between revenue and fairness via the cost-effectiveness analysis. Specifically, we set $\beta_j$ as the $\{0.1, 0.2, 0.5, 1.0, 2.0\}$ times of the average price of all items from seller $j$. Because our algorithm contains a learning part, we need counterfactual click-through rates that are not revealed in the dataset. 

For the propitiatory data set, as it includes many user/item features and to evaluate our algorithm using offline data, we follow the common practice of creating a simulator based on a deep neural network-based model \cite{chapelle2011empirical}. For the EVS dataset, we obtain experimental distribution for each of the brands. Then, for each data set, we use a simulator that generates the online environment based on these models and run the algorithm against the simulator. We also calibrate our simulator to match the distribution of clicks and purchase distribution in the data. As a robustness check, we also implement another offline evaluation technique for bandits, ``replay'' \cite{li2011unbiased}, and the final results are similar.

\subsection{Results} 

In the following,  we present the relative change of outcomes, e.g., collected GMV, only for the purpose of protecting the business data and for clarity. 

We test the performance of our proposed algorithm with an exploration step and a fairness regularizer. The benchmark algorithm for proprietary data sets is a sophisticated online learning algorithm that does not have a fairness regularizer. For the EVS data set, the benchmark greedily chooses the recommendation set and learns the CVRs with an $\epsilon$-greedy exploration scheme and without a fairness regularizer.

The results of both experiments for both datasets are given in Table \ref{tab2}. Notice that, in the table, we report the relative percentage change in outcomes lifted by Algorithm \ref{alg1} w.r.t. benchmark algorithms. Table \ref{tab2} shows that with the increasing value of the $\beta$ parameter, we observe that the number of sellers reaching the target outcomes significantly increases for larger $\beta$ for both tests. For the test on proprietary data set, one can see that when we set a mild value of $\beta$, i.e., when non-monetary value $\beta$ is 10\% of monetary revenue, our algorithm can lift the target click performance (the fairness measure on the click outcome) by 11\% and increase the revenue by 0.13\% at the same time. Similar results are obtained for the EVS dataset. This means that in some cases, our method can successfully uncover those valuable items and increase the total collected GMV for all values of $\beta/p$.

\begin{figure}[ht]
\centering
     \includegraphics[width=0.85\linewidth]{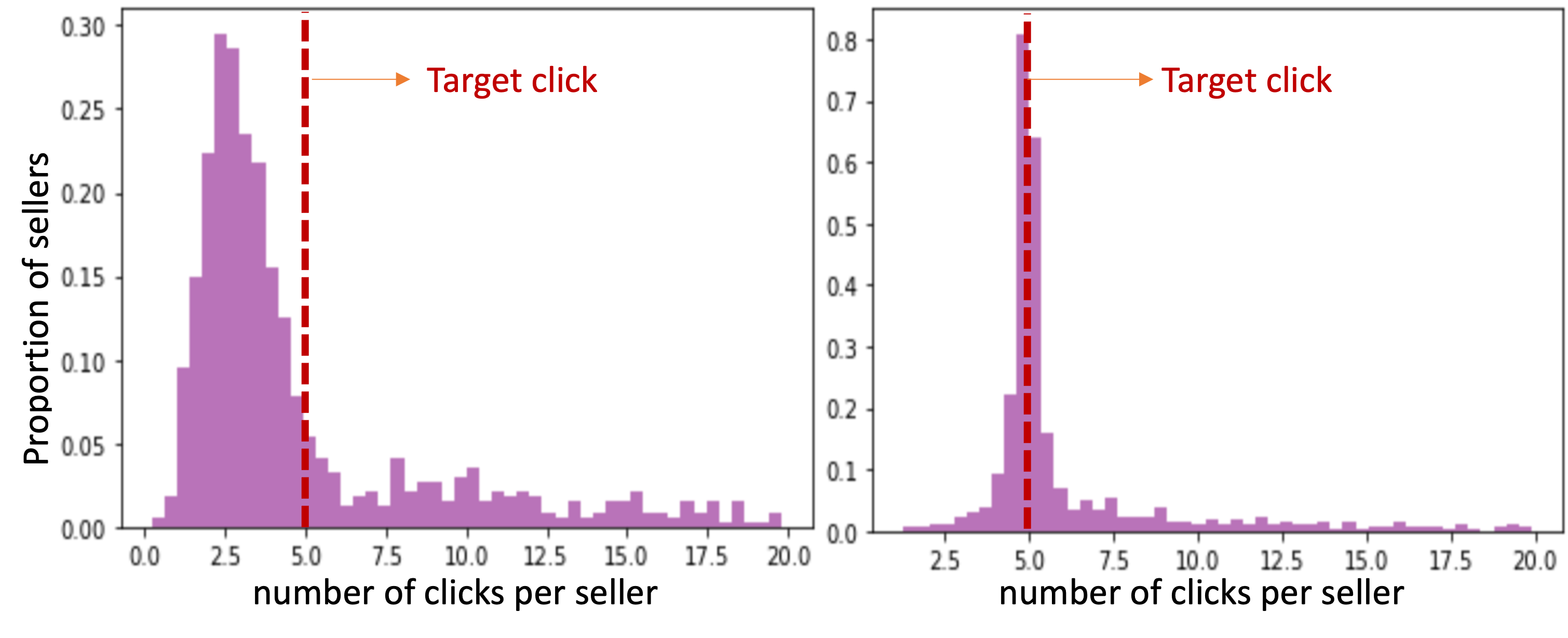}
     \caption{Baseline Performance (Left) and Algorithm Performance (Right)}\label{fig:click}
\end{figure}

To better present the importance of the fairness regularizer, we performed the test on proprietary data by running Algorithm \ref{alg1} and benchmark without exploration step and by assuming both of them know the correct underlying model. We also show the distribution of accumulated clicks over $T$ visits aggregated at the seller level in Figure \ref{fig:click}. The left subfigure shows the distribution of total clicks per $T$ visits per seller under the test done with the current benchmark, while the right subfigure shows the performance after implementing the algorithm with $\beta/p=1$, indicating a 59\% relative increase of the number of sellers that reach the target $\alpha=5$ clicks. This significant increase in outcome performance can satisfy sellers and win their loyalty, which benefits the platform in the long term.

\begin{table}[t]\centering
\caption{Relative Influence of Algorithm \ref{alg1}}
\begin{tabular}{ccccc}
\\[-1.8ex]\hline 
\hline\\[-1.8ex]
\multicolumn{5}{c}{Panel A: Relative Performance in Proprietary Dataset}\\ \hline \\[-1.8ex] 
{{$\beta/p$}} & 0.1 & 0.2 & 0.5 & 1.0 \\ \hline \\[-1.8ex] 
{GMV (Revenue) - Relative change}    & { 0.13\%} & {-0.21\%} & {-0.46\%} & {-0.84\%}  \\ 
{\# of sellers with \textgreater{}5 clicks per $T$ sessions}      & { 11\%}  & {17\%}  & {23\%}  & {28\%} \\ \hline 
\hline \\[-1.8ex] 
\multicolumn{5}{c}{Panel B: Performance in EVS Dataset}\\ \hline \\[-1.8ex] 
{{$\beta/p$}} & 0.1 & 0.2 & 0.5 & 1.0 \\ \hline \\[-1.8ex] 
{GMV (Revenue) - Relative change}   & 0.00{\%}   & -0.04{\%} & -3.71{\%} & -8.73{\%} \\ 
{\# of brands achieving the target - Alg\ref{alg1} }  & 29 & 32 & 36  & 35  \\ 
{Mean \# of brands achieving the target - benchmark}  & 16.4  & 16.4 & 16.4  & 16.4   \\ 
\hline
\hline \\[-1.8ex]  
\end{tabular}
\label{tab2}
\end{table}

\section{Conclusion and Future Research}
In this paper, we address the critical issue of seller-side fairness in online marketplaces. We introduce a notion of seller-side outcome fairness and design a gradient-based data-driven algorithm to achieve outcome fairness. Our extensive numerical experiments performed on real e-commerce data sets substantiate the efficacy of our proposed algorithm. We observed a considerable uplift in the seller fairness measures without detriment to the GMV and CTR metrics. This research, therefore, paves the way for a more equitable online marketplace, promoting diversity and potential revenue enhancement. Our findings open up new avenues for further research in this area, such as investigating other fairness measures or integrating this algorithm within larger recommendation systems. While our algorithm is effective, it is essential to continually refine these measures, as fairness in online marketplaces is a dynamic concept that evolves over time.

For future work, it will be interesting to check how our proposed algorithmic framework performs in online AB-test experiments and compares the average CTR, GMV rate versus the gain on seller fairness metrics. From the algorithmic standpoint, an interesting avenue for future research lies in deriving the theoretical upper bound of regret. This endeavor is particularly relevant given our exploration scheme, which represents a novel approach in the existing literature, grounded in the use of a ML prediction model. A deeper understanding of the regret's upper limit would not only solidify the theoretical foundations of our method but also enhance the practical applicability in dynamic and uncertain environments. Furthermore, another research direction to pursue involves addressing the challenges posed by a highly biased ML model, particularly when an optimal model cannot be realized within our searched function space. Investigating strategies to augment the robustness of our algorithm is essential. This could include developing methods to detect and correct biases in the ML model or exploring alternative model structures that are more resilient to such limitations. All of these research paths promise to significantly advance the efficacy and reliability of our algorithm in complex, real-world applications.


\bibliography{main}

\end{document}